\documentclass{easychair}

\usepackage{graphicx}
\usepackage{amsmath,amssymb,amsfonts}
\usepackage[para]{threeparttable}
\usepackage{booktabs}
\usepackage{xcolor}
\usepackage{url}
\usepackage{caption}
\usepackage{cite}

\usepackage{booktabs,multirow}
\usepackage[dvipsnames,table]{xcolor}
\usepackage{siunitx}
\definecolor{Nrow}{RGB}{255,226,220}  

\title{Results of the 2024 CommonRoad Motion Planning Competition for Autonomous Vehicles}

\author{
    Yanliang Huang\inst{1}
\and
    Xia Yan\inst{2}
\and
    Peiran Yin\inst{3}
\and
    Zhenduo Zhang\inst{4}
\and
    Zeyan Shao\inst{4}
\and
    Youran Wang\inst{1}
\and    
    Haoliang Huang\inst{1}
\and
    Matthias Althoff\inst{1}
}

\institute{
  Technical University of Munich, 
  Munich, Germany\\
  \email{\{yanliang.huang, youran.wang, althoff\}@tum.de}
 \and 
  RWTH Aachen, 
  Aachen, Germany\\
  \email{xia.yan@rwth-aachen.de}
   \and 
  Harbin Institute of Technology, 
  Harbin, China\\
  \email{23s104099@stu.hit.edu.cn}
   \and 
  Karlsruhe Institute of Technology,
  Karlsruhe, Germany\\
  \email{\{ufurn, udgto\}@student.kit.edu}
}

\authorrunning{}
\date{\phantom{date}\\}
\titlerunning{CommonRoad Motion Planning Competition 2024}

\DeclareMathOperator*{\argmax}{arg\,max}

\begin{document}

\maketitle

\begin{abstract}
Over the past decade, a wide range of motion planning approaches for autonomous vehicles has been developed to handle increasingly complex traffic scenarios. 
However, these approaches are rarely compared on standardized benchmarks, limiting the assessment of relative strengths and weaknesses. 
To address this gap, we present the setup and results of the 4th CommonRoad Motion Planning Competition held in 2024, conducted using the CommonRoad benchmark suite. This annual competition provides an open-source and reproducible framework for benchmarking motion planning algorithms.
The benchmark scenarios span highway and urban environments with diverse traffic participants, including passenger cars, buses, and bicycles. 
Planner performance is evaluated along four dimensions: efficiency, safety, comfort, and compliance with selected traffic rules. 
This report introduces the competition format and provides a comparison of representative high-performing planners from the 2023 and 2024 editions.

\end{abstract}
\section{Introduction}
\label{sec:intro}
The CommonRoad Motion Planning Competition was launched in 2021 to bring together researchers working on motion planning for autonomous vehicles. 
This report summarizes the results of the 2024 edition. The primary goal of the competition is to enable a fair and reproducible comparison of different motion planning approaches on a large number of realistic traffic scenarios. To this end, all motion planners from the participants are executed on the same hardware and evaluated in the same traffic scenarios. The benchmark scenarios are realistic since real road networks are used and the behaviors of the traffic participants are either adapted from real-world recordings or simulated by state-of-the-art traffic simulators \cite{Althoff2017a, Krajzewicz2012}. Finally, the competition additionally considers a realistic vehicle model with nonlinear dynamics and parameters taken from a real Ford Escort.
In addition to presenting the results of the competition, this report also contains a short description of the motion planners submitted by the participants, which provides insights into the different motion planning strategies.

The remainder of this report is organized as follows: First, the format of the competition is described in Sec.~\ref{sec:format} and the rules for performance evaluation are presented in Sec.~\ref{sec:evaluation}. Afterward, a description of representative participating motion planners is provided in Sec.~\ref{sec:participants}, before presenting a comparative analysis of representative competition solutions in Sec.~\ref{sec:results}.

\section{Format of the Competition}
\label{sec:format}
\begin{figure}
    \centering
    \includegraphics[width=0.7\linewidth]{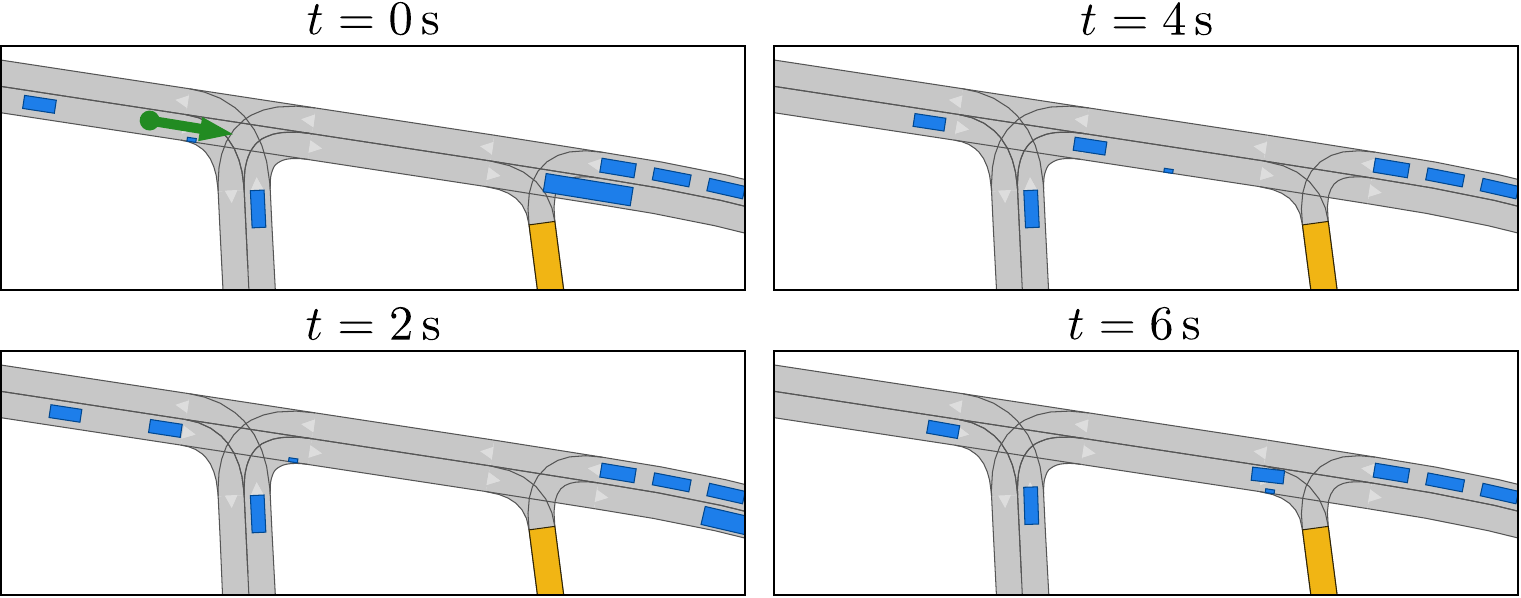}
    \caption{Example for a motion planning problem defined by the CommonRoad scenario \textit{DEU\_Flensburg-73\_1\_T-1}, where the goal set is depicted in yellow, the initial state of the ego vehicle in green, and the other traffic participants in blue.}
    \label{fig:commonRoadScenario}
\end{figure}

Teams participating in the competition are tasked with solving motion planning problems for autonomous vehicles.
An illustrative benchmark scenario is shown in Fig.~\ref{fig:commonRoadScenario}. 
The benchmarks cover both highway and urban environments and include diverse traffic participants, such as passenger cars, buses, and bicycles. 
Participants must implement motion planners that generate feasible trajectories driving the ego vehicle to a predefined goal region. 
In the 2024 edition, more than $500$ scenarios were provided, including a mix of interactive and non-interactive settings to capture a broad range of autonomous driving challenges.

The benchmark problems are specified in the CommonRoad format~\cite{Althoff2017a}, 
and participants submit Docker images of their motion planners, 
which are executed on our evaluation servers. 
Apart from the number of solved planning problems, submissions are evaluated considering efficiency, safety, comfort, and compliance with a selection of traffic rules. The exact evaluation criteria have been decided by an independent jury and are presented in detail in Sec.~\ref{sec:evaluation}. To support teams without a large software framework for motion planning, we provide useful tools for motion planning within the CommonRoad framework, such as a drivability checker \cite{PekIV20}, a route planner~\cite{commonroad_velocity_planner}, a criticality measurement toolbox \cite{lin2023crime}, etc. The competition provides two types of problems:
\begin{itemize}
    \item \textbf{Non-interactive} scenarios in which other traffic participants do not react to the behavior of the ego vehicle (with provided predictions).
    \item \textbf{Interactive} scenarios using the SUMO \cite{Krajzewicz2012, Klischat2019a} traffic simulator, in which other traffic participants react to the behavior of the ego vehicle.
\end{itemize}

To lower the entry barrier for the competition, the competition is divided into two phases:
\begin{itemize}
  \item \textbf{Phase I} is for didactic purposes (results are not considered for the final ranking), where the participants can familiarize themselves with the CommonRoad framework using public, known scenarios (a mix of non-interactive and interactive).
  \item \textbf{Phase II}: The evaluation is solely performed in Phase II using non-public, unknown, and interactive scenarios. The participants provide Docker images of their motion planners.
\end{itemize}
At the end of the competition, the results are typically presented in our annual workshop.

\section{Evaluation}
\label{sec:evaluation}
In this section, we present the evaluation criteria of the benchmark solutions, which were decided by the jury chair of the competition. We first verify if a planned trajectory is feasible (see Sec. \ref{sec:feasible_trajectories}) and reaches the goal region of the benchmark scenario. The feasible trajectories reaching the goal region are then evaluated by a cost function (see Sec. \ref{sec:cost_function}).

\subsection{Feasible Trajectories}
\label{sec:feasible_trajectories}
A \emph{trajectory} is a discrete-time sequence of ego vehicle states over the planning horizon. 
The state vector of the kinematic single-track model \cite[Sec. 2.2]{rajamani2006vehicle} is 
\( [x, y, v, \varphi, \delta]^\top\), where \(x,y\) provide the Cartesian position, 
\(v\) is the velocity, \(\varphi\) is the heading angle, 
and \(\delta\) is the steering angle. 
A trajectory is verified as feasible if it fulfills the following conditions:

\begin{itemize}
\item \textbf{Collision-free}: The occupancy of the ego vehicle does not intersect with other obstacles within the planning horizon.
\item \textbf{Kinematically feasible}: Trajectories must be feasible regarding the dynamics of a given vehicle model\footnote{\url{https://gitlab.lrz.de/tum-cps/commonroad-vehicle-models}}. In this competition, a kinematic single-track model configured by the parameters of a real vehicle is employed.
\item \textbf{Road-compliance}: The ego vehicle must stay within the road network and must not occupy walkways and bicycle paths. 
\end{itemize}
We perform the evaluation of the three conditions by the CommonRoad drivability checker \cite{PekIV20}.

\subsection{Cost Function}
\label{sec:cost_function}
We evaluate the optimality of feasible trajectories by the following cost function:
\begin{equation}
    \label{eq:cost}
    J_\mathrm{ego}=\boldsymbol{w}\,[J^{\mathrm{lon}}_J,~J_\mathrm{SR},~J_\mathrm{D},~J_\mathrm{LC}]^\top,
\end{equation}
where the individual cost terms are defined as follows:
\begin{itemize}
    \item \textbf{Jerk}: $ J^{\mathrm{lon}}_{J} = \int_{t_0}^{t_f} \dddot{s}(t) ^2 \, \mathrm{d}t$, using longitudinal jerk $\dddot{s}(t) $.
    \item \textbf{Steering rate}: $J_\mathrm{SR} = \int_{t_0}^{t_f} v_{\delta}(t)^2 \, \mathrm{d}t$, using steering velocity $v_{\delta}(t)$.
    \item \textbf{Lane center offset}: $J_\mathrm{LC} = \int_{t_0}^{t_f} d(t)^2 \, \mathrm{d}t$, where $d(t)$ is the distance to the lane center.
    \item \textbf{Distance to obstacles}: $J_\mathrm{D} = \int_{t_0}^{t_f} \max(\xi_1, \ldots, \xi_o) \, \mathrm{d}t$, where $o$ is the number of surrounding obstacles in front of the ego vehicle, $\xi_i = e^{-w_\mathtt{dist} d_i}$, $d_i$ is the distance of the ego vehicle to an obstacle, and $w_\mathtt{dist}$ is an additional required weight.
\end{itemize}
The weighting factors are $\boldsymbol{w}=[0.01,~22,~8,~5]$ and $w_\mathtt{dist}=0.2$. The cost function~(\ref{eq:cost}) is implemented in the drivability checker, and has the ID TR1. 

\section{Winners}
\label{sec:participants}
The winners of the 2024 CommonRoad Motion Planning Competition for Autonomous Vehicles are Yanliang Huang from the Technical University of Munich, Xia Yan from the RWTH Aachen, Peiran Yin from the Harbin Institute of Technology, and Zhenduo Zhang and Zeyan Shao, both from the Karlsruhe Institute of Technology. A description of their motion planner---a modified reactive planner with game-theoretic decision making---is provided in this section.

The team developed a sampling-based planner that integrates trajectory optimization with dynamic adaptation, as illustrated in Fig.~\ref{fig:structure}. 
The framework employs two complementary planning modules, each specialized for distinct scenario types: 
a \emph{modified reactive planner} for general road environments and a \emph{level-$k$ game-theoretic planner} for unsignalized intersections. 
This combination enables both broad coverage of traffic scenarios and interaction-aware decision making:
\begin{figure}[htbp]
    \centering
    \makebox[\textwidth][c]{\includegraphics[width=1\textwidth]{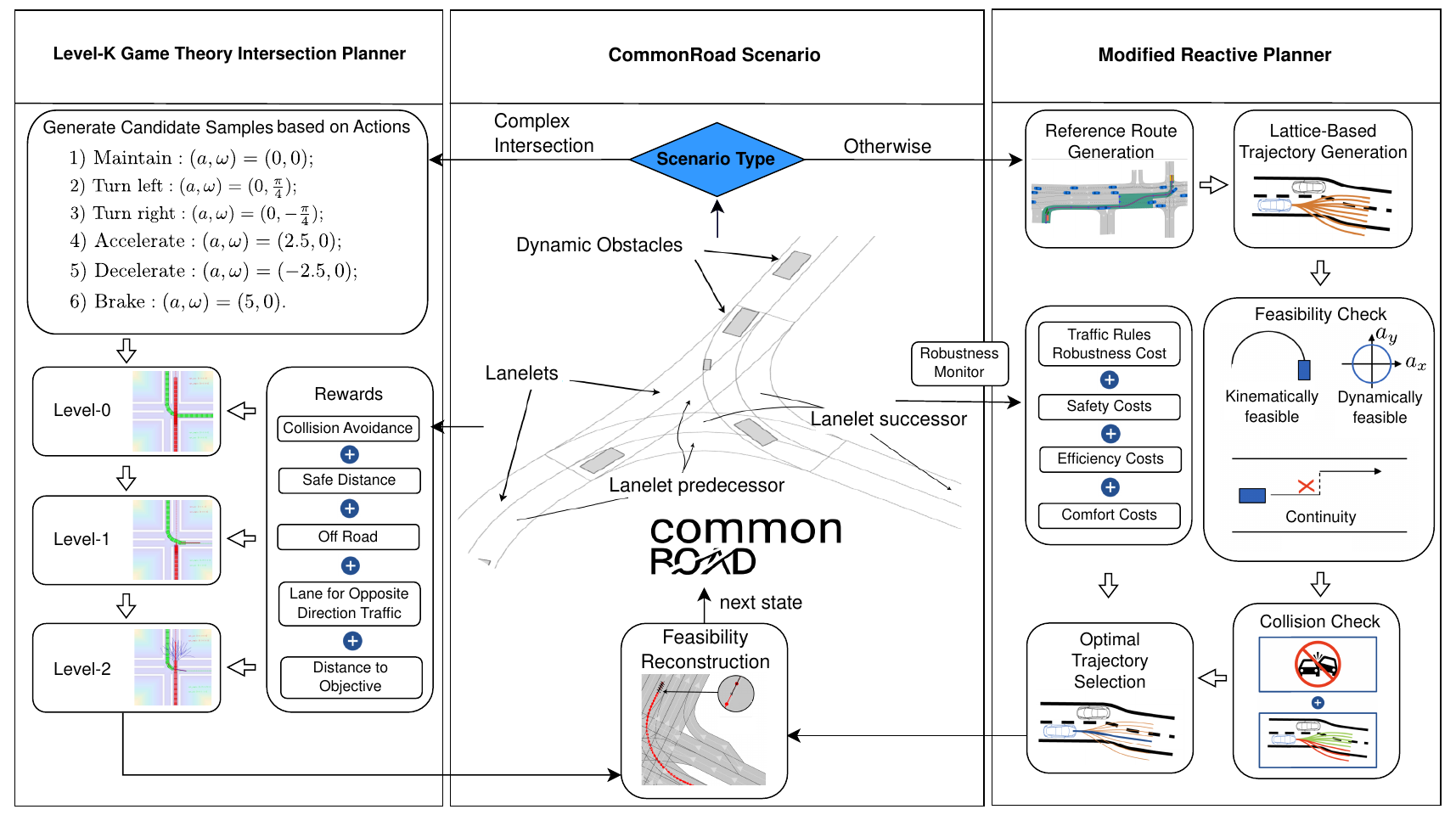}}
    \caption{Workflow of the motion planner.}
    \label{fig:structure}
\end{figure}
\subsection{Prediction Model}
In our framework, trajectories of other vehicles are predicted by forward simulation. A simplified discrete-time kinematic model is employed to propagate the state of each dynamic obstacle over a 3-second horizon, as it provides a good trade-off between computational efficiency and prediction accuracy for short-term motion forecasting:

\begin{align*}
x(t+1) &= x(t) + v(t)\cos(\theta(t))\Delta t, \\
y(t+1) &= y(t) + v(t)\sin(\theta(t))\Delta t, \\
v(t+1) &= v(t) + a(t)\Delta t, \\
\theta(t+1) &= \theta(t) + \omega(t)\Delta t,
\end{align*}
where $(x,y)$ denotes the position in Cartesian coordinates, $v$ is the longitudinal velocity, $\theta$ is the heading angle, $a$ denotes the longitudinal acceleration, and $\omega$ is the yaw rate. This simplified prediction model assumes constant control inputs $(a,\omega)$ in between discrete time steps with a step size of $\Delta t$.

\subsection{Modified Reactive Planner}

The planning approach is formulated in the Frenét frame~\cite{werling_optimal_2010}, where vehicle motions are expressed relative to a reference trajectory. This reference trajectory is obtained from the CommonRoad route planner~\cite{commonroad_velocity_planner} and connects the initial vehicle state to the goal state. To optimize the performance, the reference trajectory is adapted continuously. When lane changes occur, the route is updated based on the new lane of the ego vehicle. This updating mechanism is temporarily disabled near complex intersections, because multiple lane centerlines physically overlap or run in very close proximity there, making the nearest-lane association non-unique.

Subsequently, the CommonRoad reactive planner\footnote{\url{https://github.com/CommonRoad/commonroad-reactive-planner}} is called to generate multiple candidate trajectories in a temporal-spatial lattice using quintic curves for lateral and longitudinal sampling~\cite{werling_optimal_2010,werling_optimal_2012}. The planner evaluates each candidate trajectory for kinematic feasibility and collision avoidance, ultimately selecting the trajectory that minimizes the associated cost function.
During operation, a 3-second planning horizon, which is identical to the prediction horizon, is employed, and the replanning interval is initiated every 0.3 seconds. This interval is dynamically decreased when predefined high-risk scenarios are detected, such as high road curvature, close proximity to other vehicles, or approaching an intersection, to ensure timely plan adjustments.

The cost function follows the specification of the CommonRoad benchmarks~\cite{Althoff2017a} with an additional term to account for traffic rule compliance. The degree of rule compliance is quantified through online robustness monitors~\cite{nickovic_rtamt_2020} based on signal temporal logic (STL) for rules \textit{R\_G1} and \textit{R\_G4}~\cite{maierhofer_formalization_2020} since they are important for maintaining both safety and efficiency in interstate and urban driving scenarios:

\begin{itemize}
\item \textbf{R\_G1 - Safe distance:} The ego vehicle must maintain a safe distance from any preceding vehicle in the same lane to prevent potential collisions during sudden stops. In the event of a cut-in maneuver that violates this safe distance requirement, the ego vehicle must reestablish the appropriate safe distance within a predefined threshold $t_c$.
\item \textbf{R\_G4 - Traffic flow:} The ego vehicle must not impede traffic flow by traveling at unnecessarily low speeds. In the absence of a leading vehicle, the ego vehicle should maintain either its maximum permitted velocity or the recommended speed for the given road type.
\end{itemize}

To improve safety and adapt to varying road conditions, additional features---such as dynamic speed-limit recognition and compliance, as well as curvature-based velocity control---are integrated into the constraints of the reactive planner. 

\subsection{Level-\emph{k} Game Theory Planner for Intersections}

In unsignalized intersections, vehicles must infer and anticipate the intentions of other agents without traffic signals. These situations involve highly coupled interactions among multiple agents, where the decision of each vehicle directly influences those of others. 
While alternative interaction models (e.g., rule-based or learning-based approaches) are conceivable, the level-\emph{k} dynamic game-theoretic framework~\cite{li_game_2018} is chosen for its flexibility in representing agents with different reasoning depths and its interpretable structure, which enables explainable interaction modeling at intersections.
Our method integrates three levels of interactive reasoning (level-0, level-1, and level-2), where higher-level agents recursively reason about the predicted actions of lower-level ones. Level-$k$ denotes the depth of an agent’s recursive reasoning. At level 0, the ego vehicle acts myopically, selecting an action sequence \(\gamma \in \Gamma\)  solely to pursue its own goal without accounting for interactions with other agents. 

A discrete action space is defined as shown in Tab.~\ref{tab:maneuvers}, where each driving maneuver is characterized by a unique set of inputs. Formally, let 
\[
\gamma = ( u(0), u(1), \ldots, u(T-1) )
\] 
denote a finite sequence of discrete actions selected from the maneuver set in Tab.~\ref{tab:maneuvers}, 
where \(T\) is the prediction horizon and each \(u(t) \in \mathcal{U} \) denotes the input vector applied at time step \(t\). 
The collection of all possible action sequences is denoted by \(\Gamma\), i.e.,
\[
\Gamma = \left\{ \gamma \;\middle|\; u(t) \in \mathcal{U}, \; t = 0,\ldots,T-1 \right\},
\]
where \(\mathcal{U}\) denotes the discrete action space of maneuvers.

\begin{table}[t]
\centering
\caption{Available driving maneuvers and corresponding control inputs.}
\label{tab:maneuvers}
\begin{tabular}{c l c c}
\toprule
\# & \textbf{Description} & \textbf{Acceleration} [m/s$^2$] & \textbf{Yaw rate} [rad/s] \\
\midrule
1  & Maintain             &  0.00  &   0.00      \\
2  & Small deceleration       & -1.50  &   0.00      \\
3  & Small acceleration       &  1.50  &   0.00      \\
4  & Medium deceleration            & -3.50  &   0.00      \\
5  & Large acceleration      &  2.50  &   0.00      \\
6  & Large deceleration           & -5.00  &   0.00      \\
7  & Small steering to the left       &  0.00  &  $\pi/4$    \\
8  & Small steering to the right      &  0.00  & $-\pi/4$    \\
9  & Large steering to the left      &  0.00  &  $\pi/2$    \\
10 & Large steering to the right     &  0.00  & $-\pi/2$    \\
11 & Acceleration and steering to the left    &  1.50  &  $\pi/4$    \\
12 & Acceleration and steering to the right   &  1.50  & $-\pi/4$    \\
13 & Braking and steering to the left    & -1.50  &  $\pi/4$    \\
14 & Braking and steering to the right   & -1.50  & $-\pi/4$    \\
\bottomrule
\end{tabular}
\end{table}

At level 0, the ego vehicle chooses the action sequence that  maximizes a predefined reward function over the prediction horizon \(T\):
\begin{equation}
R(\gamma) = w_1\hat{c}(\gamma) + w_2\hat{s}(\gamma) + w_3\hat{o}(\gamma) + w_4\hat{l}(\gamma) + w_5\hat{d}(\gamma),
\end{equation}
where $\hat{c}(\gamma)$ evaluates collision risks with other vehicles, $\hat{s}(\gamma)$ ensures safe distances between vehicles, 
$\hat{o}(\gamma)$ penalizes off-road driving, $\hat{l}(\gamma)$ enforces lane keeping and proper driving direction, 
and $\hat{d}(\gamma)$ measures the distance to the target lane along the action sequence $\gamma$.
Furthermore, a level-1 agent assumes that all other agents behave as level-0 agents, i.e., they act instinctively without considering interactions. Based on this assumption, the level-1 agent optimizes its own decision accordingly. Similarly, a level-\emph{k} agent assumes that all the other agents are level-(\emph{k}-1) and predicts their behaviors under this assumption~\cite{li_game_2018}.

The decision-making process of the motion planner is based on a belief-based interactive model. For each surrounding agent, the ego vehicle maintains a belief distribution $P(k)$ representing the probability that the other agent reasons at level-\emph{k}, where $k \in \{0,1,2\}$, and $P(0) + P(1) + P(2) = 1$. The expected cumulative reward for an action sequence $\gamma$ is computed as:

\begin{equation}
R_P(\gamma) = \sum_{k=0}^2 P(k)R(\gamma|\hat{\gamma}^{other}_k),
\end{equation}
where $R(\gamma|\hat{\gamma}^{other}_k)$ represents the reward when the other agent acts according to a level-\emph{k} strategy. The ego vehicle then selects the optimal action sequence:

\begin{equation}
\hat{\gamma}^{ego}_P = \argmax_{\gamma \in \Gamma} R_P(\gamma).
\end{equation}
In case several solutions have the highest expected cumulative reward, the $\argmax$ operator returns the solution that has been computed last.

After executing the first action, the controller updates its beliefs by comparing the actual observed action of other agents, denoted as $\gamma^{other}$, with the predictions derived from different reasoning levels. Specifically, let $\hat{\gamma}^{other}_{(k,0)}$ denote the first action control inputs ($t=0$) predicted under the assumption that the other agent follows the level-$k$ model. The belief in model $k^*$ (i.e., the hypothesis that the agent is on the $k$th level) is updated by identifying which model best explains the observed behavior:

\begin{equation}
    k^* \in \arg \min_{k \in \{0,1,2\}} \| \gamma^{other} - \hat{\gamma}^{other}_{(k,0)} \|_1,
    \label{eq:belief_update}
\end{equation}
where $\gamma^{other}$ represents the actual observed inputs of the other agent at the current step, and $\| \cdot \|_1$ denotes the 1-norm~\cite{li_game_2018}.
Finally, the CommonRoad drivability checker~\cite{PekIV20} is utilized to verify the dynamic feasibility of trajectories at each time step. Notably, an emergency-braking mechanism is available as a last resort.

\section{Results}
\label{sec:results}
\subsection{Evaluation Setup}
All submitted motion planners were evaluated on a machine with an Intel Core i7-14700KF (base 3.4\,GHz, boost up to 5.6\,GHz). The evaluation is timed out after 6 hours, and at most two cores can be used. Note that the time limit penalizes computation time since motion planners that are faster will be able to solve more benchmarks within the given time frame.

To enable a cross-year comparison, we evaluate the 2024 winning planner (TUM-2024; the team of \emph{Huang, Yanliang}) against the winning planner in 2023 from Stony Brook University (SB-2023, the team of \emph{Kochdumper, Niklas})~\cite{kochdumper2023results} on the same interactive CommonRoad benchmarks.

\subsection{Benchmark Coverage Comparison}
\begin{figure}[htbp]
    \centering
    \includegraphics[width=0.5\textwidth]{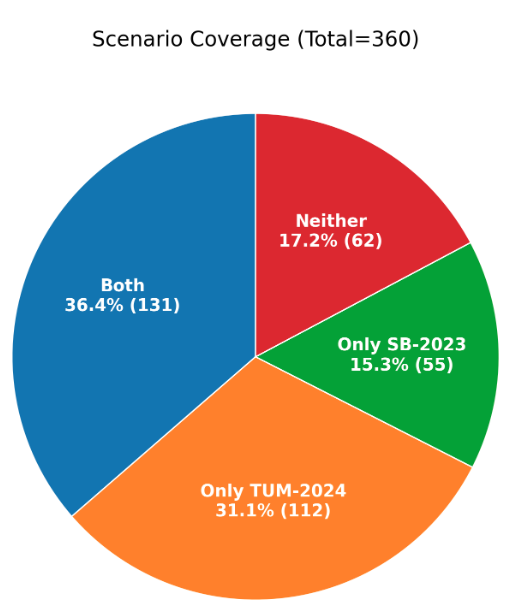}
    \caption{Scenario coverage of the two motion planners: TUM-2024 and SB-2023.}
    \label{fig:coverage_pie}
\end{figure}

Figure~\ref{fig:coverage_pie} summarizes benchmark coverage. Out of 360 benchmarks, 131  (36.4\%) scenarios are solved by both planners. TUM-2024 additionally solved 112 (31.1\%) scenarios that SB-2023 did not, while SB-2023 uniquely solved 55 (15.3\%); 62 scenarios (17.2\%) remained unsolved by both. This indicates a clear scenario-coverage advantage for TUM-2024 under the two-core, 6-hour budget.
The resulting planner demonstrates robust performance across a wide range of scenarios, effectively balancing safety, efficiency, and passenger comfort while reliably reaching the designated goal state.
\subsection{Map-wise Performance Comparison}

Tab.~\ref{tab:city-overall} and \ref{tab:city-common} report results grouped by road map clusters. 
The motivation of this comparison is that the difference in road structure influences both coverage and solution quality.  
TUM-2024 achieved $20$--$70\%$ higher benchmark coverage across all road map clusters, 
whereas SB-2023 obtained $30$--$40\%$ lower TR1 costs in general with reduced variance and better worst-case performance.
To remove the confounding effect of differing coverage, 
Tab.~\ref{tab:city-common} reports the performance on benchmarks that are solved by \emph{both} planners. 
On this shared set, SB-2023 achieved approximately $40\%$ lower average TR1 costs across all clusters 
and also exhibited lower variance and more favorable worst-case performance, 
indicating better per-benchmark performance. The comparatively lower trajectory quality of TUM-2024 can be attributed to its simplified prediction model of other agents, 
where inaccurate forecasts may cause the planner to generate overly conservative trajectories.

In conclusion, the comparison results highlight a clear trade-off. 
TUM-2024 emphasizes breadth, achieving substantially higher coverage under a fixed two-core, six-hour budget. 
In contrast, SB-2023 emphasizes quality: on benchmarks where both succeed, it yields lower TR1 costs (i.e., achieves better performance in terms of comfort, safety, and rule compliance). 
Thus, TUM-2024 is favored when evaluation emphasizes the number of solved benchmarks, 
while SB-2023 dominates when the focus is on solution quality.

\begin{table}[t]
\caption{Performance on all benchmarks.}
\captionsetup[table]{skip=6pt} 
\label{tab:city-overall}
\centering
\small
\renewcommand{\arraystretch}{1.15}
\setlength{\tabcolsep}{6pt}

\begin{tabular}{
  l
  c  
  l  
  S[table-format=3.0, round-precision=0] 
  S[table-format=2.3, round-mode=places, round-precision=3]   
  S[table-format=2.3, round-mode=places, round-precision=3]   
  S[table-format=2.3, round-mode=places, round-precision=3]   
}
\toprule
\textbf{City} & \textbf{Total} & \textbf{Competitor} &
\textbf{Overall Solved} & \textbf{Mean} & \textbf{Std} & \textbf{Worst} \\
\midrule

\multirow{2}{*}{\cellcolor{white}Aachen}
  & \multirow{2}{*}{\num{120}} & TUM-2024
  & 82 & 7.629 & 9.339 & 66.384 \\
& & \cellcolor{Nrow}\textbf{SB-2023}
  & \cellcolor{Nrow}68
  & \cellcolor{Nrow}{\bfseries 4.622}
  & \cellcolor{Nrow}7.811
  & \cellcolor{Nrow}45.203 \\
\midrule

\multirow{2}{*}{\cellcolor{white}Cologne}
  & \multirow{2}{*}{\num{180}} & TUM-2024
  & 117 & 5.966 & 5.699 & 35.032 \\
& & \cellcolor{Nrow}\textbf{SB-2023}
  & \cellcolor{Nrow}92
  & \cellcolor{Nrow}{\bfseries 3.972}
  & \cellcolor{Nrow}4.766
  & \cellcolor{Nrow}16.495 \\
\midrule

\multirow{2}{*}{\cellcolor{white}Dresden}
  & \multirow{2}{*}{\num{60}} & TUM-2024
  & 44 & 7.059 & 7.321 & 30.841 \\
& & \cellcolor{Nrow}\textbf{SB-2023}
  & \cellcolor{Nrow}26
  & \cellcolor{Nrow}{\bfseries 4.061}
  & \cellcolor{Nrow}4.049
  & \cellcolor{Nrow}14.142 \\
\bottomrule
\end{tabular}
\end{table}

\begin{table}[t]
\caption{Performance on benchmarks solved by both TUM-2024 and SB-2023.}
\label{tab:city-common}
\centering
\small
\renewcommand{\arraystretch}{1.15}
\setlength{\tabcolsep}{6pt}

\begin{tabular}{
  l
  c  
  c  
  l
  S[table-format=2.3, round-mode=places, round-precision=3]   
  S[table-format=2.3, round-mode=places, round-precision=3]   
  S[table-format=2.3, round-mode=places, round-precision=3]   
}
\toprule
\textbf{City} & \textbf{Total} & \textbf{Solved by Both} & \textbf{Competitor} &
\textbf{Mean} & \textbf{Std} & \textbf{Worst} \\
\midrule

\multirow{2}{*}{\cellcolor{white}Aachen}
  & \multirow{2}{*}{\num{120}} & \multirow{2}{*}{\num{45}} & TUM-2024
  & 7.635 & 10.980 & 66.384 \\
&  &  & \cellcolor{Nrow}\textbf{SB-2023}
  & \cellcolor{Nrow}{\bfseries 2.944}
  & \cellcolor{Nrow}5.789
  & \cellcolor{Nrow}33.300 \\
\midrule

\multirow{2}{*}{\cellcolor{white}Cologne}
  & \multirow{2}{*}{\num{180}} & \multirow{2}{*}{\num{65}} & TUM-2024
    & 6.767 & 6.595 & 35.032 \\
  &  &  & \cellcolor{Nrow}\textbf{SB-2023}
    & \cellcolor{Nrow}{\bfseries 3.267}
    & \cellcolor{Nrow}4.336
    & \cellcolor{Nrow}14.980 \\
  \midrule

\multirow{2}{*}{\cellcolor{white}Dresden}
  & \multirow{2}{*}{\num{60}} & \multirow{2}{*}{\num{21}} & TUM-2024
  & 5.784 & 6.505 & 26.749 \\
&  &  & \cellcolor{Nrow}\textbf{SB-2023}
  & \cellcolor{Nrow}{\bfseries 3.483}
  & \cellcolor{Nrow}3.117
  & \cellcolor{Nrow}10.979 \\
\bottomrule
\end{tabular}
\end{table}

\section{Conclusion}
\label{sec:conclusion}
This report summarizes the results of the 4th CommonRoad Motion Planning Competition for Autonomous Vehicles held in 2024. 
The winning planner from the Technical University of Munich (TUM-2024) employs a sampling-based strategy in the Frenét frame with model-based prediction of surrounding agents. The highlight of their approach is that decision-making at intersections is resolved by formulating level-$k$ games.
We compared its performance to the award-winning 2023 planner from Stony Brook University (SB-2023), which combines optimization-based planning with reachability analysis. 

TUM-2024 achieved substantially higher scenario coverage within the fixed compute budget, 
whereas SB-2023 consistently produced lower TR1 costs on the set of benchmarks solved by both planners. 
The results suggest that TUM-2024 could be further improved by integrating more accurate prediction models of other agents, 
thereby reducing overly conservative behavior while retaining its scalability advantage.

\bibliographystyle{abbrv}
\bibliography{bib/general.bib}

\end{document}